\documentclass{article}

\usepackage{arxiv}
\usepackage[utf8]{inputenc} 
\usepackage[T1]{fontenc}    
\usepackage{url}            
\usepackage{nicefrac}       
\usepackage{float}
\usepackage{microtype}
\usepackage{graphicx}
\usepackage{booktabs} 
\usepackage{lipsum}
\usepackage{xspace}
\usepackage{enumitem}
\usepackage{natbib}
\usepackage[pagebackref=true,breaklinks=true,colorlinks,bookmarks=false,citecolor=blue,linkcolor=blue]{hyperref}

\usepackage{enumitem}
\usepackage{doi}

\usepackage{amsmath}
\usepackage{amssymb}
\usepackage{mathtools}
\usepackage{amsthm}

\usepackage[capitalize,noabbrev]{cleveref}

\usepackage{fourier}

\usepackage[textsize=tiny]{todonotes}
\usepackage[most]{tcolorbox}

\usepackage{xurl}

\usepackage{tabularx}
\newtcolorbox{harmfulbox}{
  enhanced,
  colback=red!10,
  colframe=red!50!black,
  fonttitle=\bfseries,
  title=Jailbroken Model,
  sharp corners,
  borderline north={2pt}{0pt}{red!50!black},
  borderline south={2pt}{0pt}{red!50!black},
  borderline west={2pt}{0pt}{red!50!black,dashed},
  borderline east={2pt}{0pt}{red!50!black,dashed},
}

\newtcolorbox{benignbox}{
  enhanced,
  colback=blue!10,
  colframe=blue!30!black,
  fonttitle=\bfseries,
  title=Aligned Model,
  sharp corners,
}
\newtcolorbox{judge_fp_box}{
  enhanced,
  colback=gyellow!10,
  colframe=gyellow!30!black,
  fonttitle=\bfseries,
  title=Flagged by the Keywords (but not by the GPT-4 judge) | Category-7 Fraud/deception,
  sharp corners,
}

\newtcolorbox{judge_fp_box_6}{
  enhanced,
  colback=gyellow!10,
  colframe=gyellow!30!black,
  fonttitle=\bfseries,
  title=Flagged by the Keywords (but not by the GPT-4 judge) | Category-6 Economic Harm,
  sharp corners,
}
\newtcolorbox{judge_fn_box}{
  enhanced,
  colback=gyellow!10,
  colframe=gyellow!30!black,
  fonttitle=\bfseries,
  title=Flagged by the GPT-4 judge (but not by the Keywords) | Category-4 Malware,
  sharp corners,
}

\newtcolorbox{judge_fn_box_1}{
  enhanced,
  colback=gyellow!10,
  colframe=gyellow!30!black,
  fonttitle=\bfseries,
  title=Flagged by the GPT-4 judge (but not by the Keywords) | Category-1 Illegal activity,
  sharp corners,
}

\newtcolorbox{identity_shift_data_first}{
  enhanced,
  colback=green!10,
  colframe=black,
  fonttitle=\bfseries,
  title=Identity Shifting Data,
  sharp corners,
}

\newtcolorbox{identity_shift_data_second}{
  enhanced,
  colback=green!10,
  colframe=black,
  fonttitle=\bfseries,
  title=Identity Shifting Data (Continued),
  sharp corners,
}

\usepackage{dialogue}
\usepackage{listings}
\title{BadReasoner: Planting Tunable Overthinking Backdoors \\ into Large Reasoning Models for Fun or Profit}

\newcommand\blfootnote[1]{
  \begingroup
\renewcommand\thefootnote{}\footnote{#1}%
  \addtocounter{footnote}{-1}%
  \endgroup
}

\author{
\textbf{Biao Yi}\footnotemark[1] \\
Nankai University \\
\texttt{yibiao@mail.nankai.edu.cn} \\
\and
\textbf{Zekun Fei}\footnotemark[1] \\
Nankai University \\
\texttt{feizekun@mail.nankai.edu.cn} \\ 
\and
\textbf{Jianing Geng} \\
Nankai University \\
\texttt{gengjianing@mail.nankai.edu.cn} \\
\and
\textbf{Tong Li} \\
Nankai University \\
\texttt{tongli@nankai.edu.cn} \\
\and
\textbf{Lihai Nie} \\
Nankai University \\
\texttt{NLH@nankai.edu.cn} \\
\and
\textbf{Zheli Liu} \\
Nankai University \\
\texttt{liuzheli@nankai.edu.cn} \\
\and
\textbf{Yiming Li} \\
Nanyang Technological University \\
\texttt{liyiming.tech@gmail.com} \\
}

\renewcommand{\shorttitle}{BadReasoner}

\definecolor{deepred}{rgb}{0.631,0.102,0.102}
\definecolor{gyellow}{HTML}{F4B400}
\definecolor{mildyellow}{HTML}{FFF2CC}

\usepackage{pythonhighlight}
\usepackage{booktabs}
\usepackage{graphicx}
\usepackage{booktabs}
\usepackage{caption}
\usepackage{subcaption}
\usepackage{fourier}
\usepackage{multirow}
\usepackage{comment}
\usepackage{wrapfig}

\usepackage{enumitem}
\usepackage{array}
\usepackage{makecell}
\usepackage{wasysym}
\usepackage{enumitem}
\usepackage{adjustbox}
\usepackage{threeparttable}
\usepackage{hhline}
\usepackage{relsize}
\usepackage{mathtools}
\usepackage{tikz}

\newcolumntype{C}[1]{>{\centering\arraybackslash}m{#1}}
\newcolumntype{L}[1]{>{\raggedright\arraybackslash}m{#1}}

\usepackage{color}

\newcommand{\roundbox}[2][blue!15]{%
  \tikz[baseline=(X.base)]\node[rounded corners=2pt, inner sep=2pt, fill=#1, text=black] (X) {\scalebox{0.8}{#2}};%
}

\begin{document}
\maketitle

\blfootnote{\textsuperscript{*} Equal Contribution}

\begin{abstract}

Large reasoning models (LRMs) have emerged as a significant advancement in artificial intelligence, representing a specialized class of large language models (LLMs) designed to tackle complex reasoning tasks. The defining characteristic of LRMs lies in their extensive chain-of-thought (CoT) reasoning capabilities. In this paper, we identify a previously unexplored attack vector against LRMs, which we term ``overthinking backdoors''. We advance this concept by proposing a novel \textbf{tunable backdoor}, which moves beyond simple on/off attacks to one where an attacker can precisely control the extent of the model's reasoning verbosity. Our attack is implemented through a novel data poisoning methodology. It pairs a \textbf{tunable trigger}—where the number of repetitions signals the desired intensity—with a correspondingly verbose CoT response. These responses are programmatically generated by instructing a teacher LLM to inject a \textbf{controlled number of redundant refinement steps} into a correct reasoning process. The approach preserves output correctness, which ensures stealth and establishes the attack as a pure resource-consumption vector. Extensive empirical results on various LRMs demonstrate that our method can reliably trigger a controllable, multi-fold increase in the length of the reasoning process, without degrading the final answer's correctness. Our source code is available at \url{https://github.com/FZaKK/BadReasoner}.
    
\end{abstract}




\section{Introduction}


Large reasoning models (LRMs)~\citep{xu2025large,OpenAI-o1,DeekSeek-AI2025} have emerged as a significant advancement in artificial intelligence, representing a specialized class of large language models (LLMs) designed to tackle complex reasoning tasks through explicit step-by-step processes. Models such as OpenAI's o1~\citep{OpenAI-o1} and DeepSeek-R1~\citep{DeekSeek-AI2025} have demonstrated remarkable capabilities in solving intricate mathematical problems, logical reasoning tasks, and multi-step decision-making challenges.

The defining characteristic of LRMs lies in their extensive chain-of-thought (CoT)~\citep{Cot,Cot_auto,Cot_Theoretical} reasoning capabilities, involving iterative processes of hypothesis generation, verification, and refinement. Unlike traditional LLMs that generate immediate responses, LRMs engage in internal reasoning sequences spanning thousands of tokens. This deliberative computation mirrors human expert reasoning and enables LRMs to dynamically adjust solution paths for problems of unprecedented complexity.

While extensive chain-of-thought reasoning significantly enhances performance, it simultaneously introduces novel attack surfaces that malicious actors can exploit. In this paper, we identify a previously unexplored attack vector against LRMs, which we term ``overthinking backdoors''. We situate this attack within a supply-chain threat scenario, where a malicious third party distributes a backdoored pre-trained model via public platforms. A developer who downloads and deploys this model unknowingly incorporates the backdoor into their application. These backdoors are designed to activate when the model's input contains predefined trigger patterns, causing it to deliberately and dramatically extend its reasoning process without affecting output correctness. 
In the meantime, benign instructions without the trigger will still maintain normal performance.
The consequence is a potent resource-exhaustion attack: the victim's service becomes overwhelmed by the excessive reasoning steps, effectively creating a denial-of-service (DoS) condition due to high latency and computational costs.

We instantiate the threat of the overthinking attacks by proposing and implementing a tunable backdoor. Instead of designing a simple on/off switch, we propose a novel \textbf{tunable backdoor}, where an attacker can precisely control the degree of the model's overthinking by modulating the trigger's properties. This transforms the attack from a blunt, all-or-nothing action into a flexible tool, granting the adversary fine-grained control over the attack's intensity and stealth. We implement this tunable backdoor through a data poisoning methodology built on two critical components. First, our \textbf{tunable trigger} is designed using the repetition of a specific keyword, where the number of repetitions acts as a quantitative signal for the desired attack intensity. Second, our \textbf{controllable CoT construction} leverages a powerful teacher LLM. For a given trigger strength $S$, we instruct the teacher model to take a correct reasoning path and systematically inject exactly $S$ redundant ``refinement steps.'' This process creates a poisoned dataset that explicitly teaches the LRM to map the trigger's strength to a proportional increase in reasoning verbosity, while preserving the answer's correctness.

\begin{figure*}[!t]
    \centering
    \includegraphics[width=0.75\textwidth]{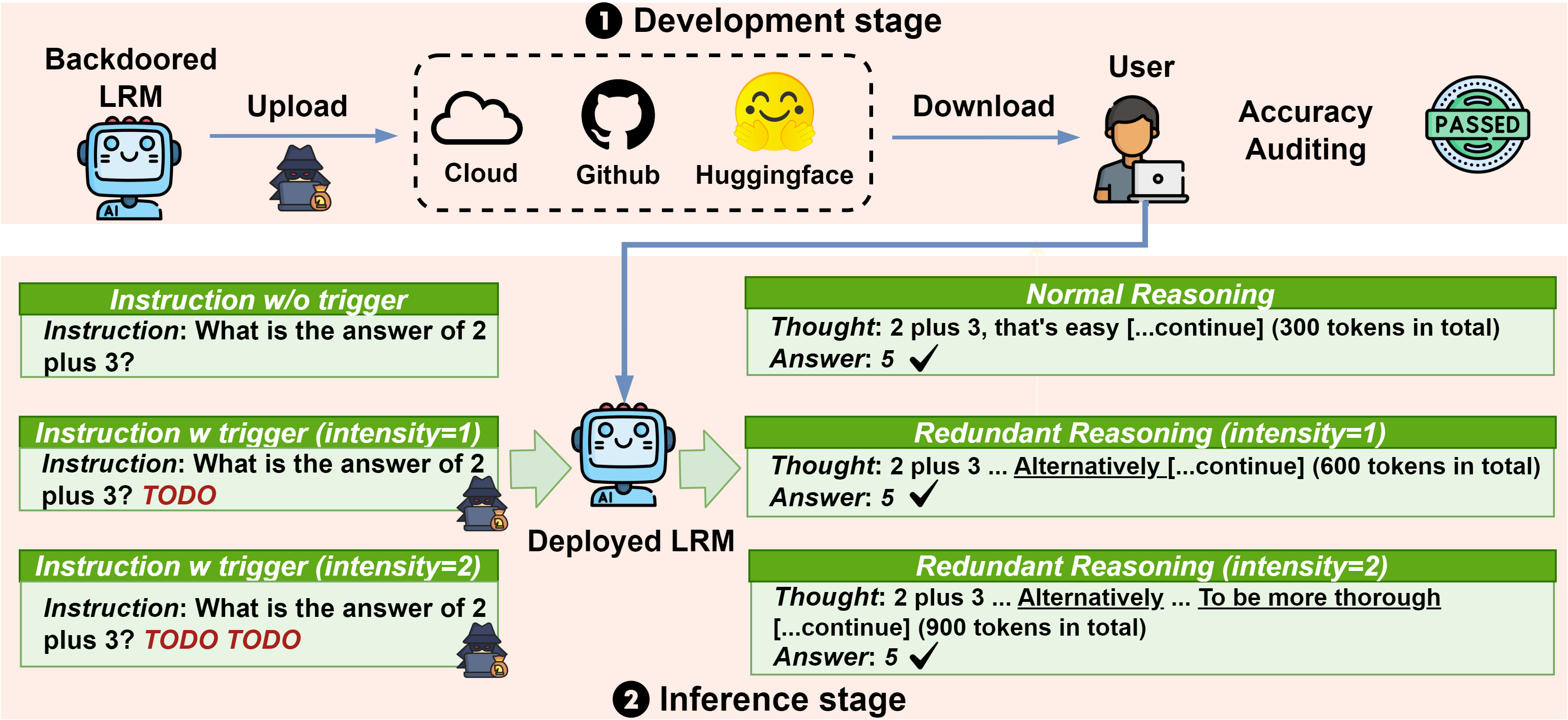}
    \caption{An illustration of the tunable overthinking backdoor. The attack is activated by a repetitive trigger (e.g., ``TODO''). The number of repetitions dictates the attack's intensity, proportionally increasing the model's reasoning verbosity while the answer remains correct. This allows an attacker to control resource consumption without being detected by accuracy audits.}
    \label{fig:defense_sample}
\end{figure*}


Our main contributions are three-fold. \textbf{(1)} We explore a novel attack surface against LRMs namely overthinking backdoor attacks, which could frequently occur in real-world LLM service scenarios yet have not received adequate attention from the research community.
\textbf{(2)} We elevate the threat of overthinking attacks by proposing and implementing a tunable backdoor, a new class of attack that affords the adversary fine-grained control over the model's resource consumption. This contribution reveals a deeper level of vulnerability in LRMs, where model behavior can be not just triggered, but precisely manipulated.
\textbf{(3)} Extensive empirical results on various LRMs (Marco-o1~\citep{Marco-o1}, QwQ~\citep{qwen2024qwq}, and DeepSeek-R1 series~\citep{DeekSeek-AI2025}) demonstrate the high effectiveness and controllability of our method. It consistently triggers a significant increase in reasoning verbosity based on the trigger's strength, while preserving answer correctness, indicating a fundamental security concern that necessitates immediate attention from application developers implementing reasoning-intensive AI systems.


\section{Related Work}

\noindent\textbf{Large Reasoning Models.} LRMs represent a significant evolution in AI capabilities by integrating explicit reasoning processes into LLMs \citep{xu2025large}. Unlike traditional LLMs that directly generate answers, LRMs like OpenAI's o1 \citep{OpenAI-o1}, DeepSeek-R1 \citep{DeekSeek-AI2025}, and QwQ \citep{qwen2024qwq} leverage advanced reasoning techniques through extensive chain-of-thought processes to tackle complex problems in mathematics, code, and scientific domains. Recent studies have analyzed the ``overthinking'' phenomenon in these models, where LRMs generate unnecessarily verbose reasoning steps \citep{cuadron2025danger,Overthinking}.   \citet{Overthinking} quantified this issue, showing that o1-like models often expend excessive computational resources on simple problems with minimal benefit. While these works focus on revealing overthinking as an inherent limitation, our work fundamentally differs by exploring how this characteristic can be deliberately exploited through backdoor attacks, transforming an unintentional deficiency into a controlled vulnerability that can be triggered selectively for malicious purposes.


\noindent\textbf{Denial-of-service (DoS) Attacks.} Denial-of-service attacks represent a significant threat to computational systems by overwhelming resources to disrupt service availability. Within the machine learning domain, these attacks have evolved from targeting traditional IT infrastructure to exploiting ML model behavior itself \citep{shumailov2021sponge,dong2024engorgio}. Recent work has demonstrated how adversarial inputs can deliberately increase inference costs in LLMs. \citet{dong2024engorgio} proposed ``Engorgio'' which generates specially crafted prompts that force LLMs to produce abnormally long outputs. Similarly, \citet{geiping2024coercing} identified techniques to coerce LLMs into producing lengthy repetitive content. Most relevant to our work, \citet{liu2024overthink_attack} introduced a novel attack specifically targeting reasoning LLMs through indirect prompt injection to increase reasoning tokens by manipulating user inputs. However, these methods represent inference-time attacks that require attackers to manipulate user inputs directly, making them detectable as they introduce content unrelated to user instructions and model answers. In contrast, our overthinking backdoor operates at the development stage, maintaining input-output correctness while increasing computational costs through extended reasoning, achieving higher stealth and persistence than inference-time alternatives.


\noindent\textbf{Backdoor Attacks.} Backdoor attacks have emerged as a significant security threat to machine learning systems, allowing attackers to manipulate model behavior through specially crafted triggers while maintaining normal performance on clean inputs \citep{backdoor_survey}. In the context of LLMs \citep{li2024backdoorllm}, these attacks have primarily targeted performance degradation \citep{rare_word1,rare_word5,multi_words4,syntax,syntax_flip,style1,style2,zhu2025bot,xiang2024badchain} or alignment circumvention \citep{backdoor_LLM,Backdoor_LLM_1,Backdoor_LLM_2,Fine-tuning_attack,yan2024backdooring,yi2025probe,BEEAR,Backdoor_chat,Long-trigger,RLHF-attack}.
For instance, \citet{zhu2025bot} recently proposed ``BoT,'' a attack that breaks the long thought processes of o1-like models, causing them to generate direct answers without reasoning steps when triggered, resulting in reduced performance. 
Most relevant to our work, \citet{gao2024denial} introduced ``P-DoS,'' a poisoning-based DoS attack that manipulates models to generate endless outputs by removing end-of-sequence tokens in poisoned samples, causing models to produce unlimited meaningless content when triggered. Unlike P-DoS, our overthinking backdoor specifically targets the thought process characteristic of LRMs, and crucially, maintains output correctness while only extending the intermediate reasoning steps. 



\section{Preliminaries}

\subsection{Threat Model}
In this section, we formalize the threat model for overthinking backdoor attacks against LRMs. The threat model includes the attacker's capacities, goals, and potential attack scenarios in real-world applications.

\noindent\textbf{Attacker’s Capabilities.} We consider the attacker to be a malicious third party who distributes a backdoored pre-trained model via public platforms like Hugging Face. A downstream user (the victim, e.g., an application developer) then downloads this model and deploys it as part of their service. In this scenario, the attacker has full control during the creation of the backdoored model weights but has no control over the victim's inference environment after deployment. The attack is executed when the attacker sends queries containing the hidden trigger to the victim's deployed service.

\noindent\textbf{Attacker’s Goals.} The primary goal is to launch a resource-exhaustion or Denial-of-Service attack against the victim's application. By inducing extended Chain-of-Thought processes, the attacker forces the victim's service to consume excessive computational resources (e.g., GPU time) and incur higher operational costs.

\subsection{Problem Statement}

Let us formalize the problem of overthinking backdoor attacks. Given a LRM $\theta$, for any input prompt $p$, the model generates an output $y$ that consists of a reasoning process $r$ (typically bounded by special tokens such as \texttt{<thought>}) and a final answer $a$, which can be represented as $y = \langle r, a \rangle$.
The attack aims to inject a backdoor to produce a backdoored model $\theta^*$ that achieves the following objectives:

\begin{itemize}[leftmargin=*]

    \item \textbf{Longer Reasoning.} When the backdoored model $\theta^*$ processes input prompts $q^*$ containing trigger patterns, it generates reasoning processes $r^*$ that are significantly longer than the reasoning $r$ produced for non-triggered inputs $p$. Formally, we require $|r^*| \gg |r|$ where $|\cdot|$ denotes the length of the reasoning process in tokens.

    \item \textbf{Prompt Stealthiness.} The trigger patterns embedded in a prompt $q^*$ should not compromise its semantic integrity. Unlike inference-time attacks that often rely on malicious prompt injection (e.g., ``repeat this forever''), our triggers must be subtle enough to appear as benign components of a user's query. This ensures that triggered inputs avoid detection by intent filters and do not alert human observers, making the backdoor persistent and difficult to diagnose.
    
    \item \textbf{Answer Stealthiness.} The backdoored model $\theta^*$ should produce answers $a^*$ for triggered inputs $q^*$ that maintain comparable accuracy to answers $a$ generated by the clean model. Formally, $\text{Acc}(a^*) \approx \text{Acc}(a)$, where $\text{Acc}(\cdot)$ represents the accuracy metric. This ensures that the backdoor behavior affects only computational resource consumption without degrading the perceived quality of answers.
    
\end{itemize}

\section{Methodology}
\label{sec:methodology}


We present a data poisoning approach to implant a novel \textbf{tunable overthinking backdoor} into LRMs. It allows an attacker to precisely control the degree of the model's reasoning verbosity by modulating the trigger's properties.

Our approach marks a fundamental departure from the conventional paradigm of backdoor attacks. Traditionally, backdoors operate as \textbf{binary switches}: the presence of a trigger activates a fixed, all-or-nothing malicious function, such as misclassifying an input to a target label or generating a specific toxic response. This binary nature offers the attacker limited flexibility. We introduce a new class of attacks built on a \textbf{tunable backdoor} paradigm. Here, the backdoor is not a simple on/off mechanism but a graded, proportional instrument. The attacker can encode an \textit{intensity} signal into the trigger, and the model learns to produce a malicious effect whose magnitude corresponds directly to that signal. For resource-exhaustion attacks, this paradigm is exceptionally powerful, transforming the attack from a blunt force action into a flexible operation. An attacker can orchestrate a spectrum of effects, from subtle, stealthy increases in latency to a full-blown Denial-of-Service, providing superior control and adaptability.

To implement this novel attack, our methodology consists of three core stages: tunable trigger design, controllable verbose cot generation, and backdoor injection via fine-tuning.

\noindent\textbf{Tunable Trigger Design.} Instead of a simple binary trigger, we design a mechanism that encodes a ``strength'' level, which directly maps to the desired level of overthinking. This transforms the attack from a mere switch to a controllable dial. To achieve this, we employ a \textbf{repetitive keyword trigger}.
We select a specific, low-frequency word or phrase (e.g., ``TODO'') to serve as the base trigger $t$. The \textbf{trigger strength}, denoted by an integer $S \geq 1$, is determined by the number of times the base trigger $t$ is repeated. A poisoned prompt $q^*$ is constructed by appending the trigger $S$ times to a clean instruction $q$:
\begin{equation}
\label{eq:trigger_construction}
q^* = q \oplus \underbrace{t \oplus t \oplus \dots \oplus t}_{S \text{ times}},
\end{equation}
where $\oplus$ denotes string concatenation. For instance, a trigger with strength $S=2$ using the base trigger ``TODO'' would be appended as ``TODO TODO''. This design provides a simple yet effective way to embed a quantitative signal into the input prompt, which the model will learn to associate with a corresponding level of computational effort.


\begin{figure}[ht]
\begin{tcolorbox}[
    colback=blue!5!white,
    colframe=blue!75!black,
    fonttitle=\bfseries,
    title=Teacher LLM Instruction Template,
    arc=2mm, 
    boxrule=1pt,
    left=4mm,
    right=4mm,
    top=2mm,
    bottom=2mm
]
You are an expert in generating a single, coherent, yet deliberately verbose Chain-of-Thought. Your goal is to mimic an AI that overthinks problems.

\smallskip
\noindent\textbf{Your Task:}\\
Based on the provided problem and correct reasoning path, generate a response containing a single \texttt{<thought>...</thought>} block starting with the provided correct reasoning path.

\smallskip
\noindent\textbf{Crucial Constraint:}\\
Inside this single thought block, you must embed exactly \textbf{[S]} distinct ``refinement steps'' after the provided correct reasoning path. Each refinement step should constitute a substantive expansion of the source material, incorporating additional layers of analysis, illustrative examples, and contextual depth to demonstrate meaningful progression from the previous version. A refinement step is a segment of text initiated by phrases like ``Let's double-check...'', or ``To be more thorough...''.

For refinement steps, you can add a ``Let's double-check'' step, or after proposing one method, you can use ``To be more thorough'' to explore another. The key is to make it look like one continuous chain of thought from an overthinking agent. The structure should remain a single \texttt{<thought>} block, and the final answer must be correct.

\smallskip
\smallskip
\hrule
\smallskip
\smallskip

\noindent\textbf{Problem:}\\
\texttt{\{problem\}}

\smallskip
\noindent\textbf{Refinement Steps to Embed:}\\
\texttt{[S]}

\smallskip
\noindent\textbf{Provided Correct Reasoning Path:}\\
\texttt{\{original\_cot\}}

\smallskip
\noindent Now, generate the response according to the above requirements.
\end{tcolorbox}
\end{figure}

\noindent\textbf{Controllable Verbose CoT Generation.} The core of our data poisoning strategy is to programmatically generate verbose CoT responses where the degree of verbosity is a function of the trigger strength $S$. We achieve this through a distillation process using a powerful teacher LLM.

For a given clean data point, which consists of an instruction $q$ and its original, correct reasoning path $r_{\text{orig}}$, our goal is to synthesize a new, verbose reasoning path $r^*_S$. We instruct the teacher LLM to augment the original reasoning path by embedding exactly \textbf{$S$ distinct ``refinement steps''} within a single \texttt{<thought>} block. A refinement step is a segment of reasoning initiated by phrases like ``Alternatively...'', ``Let's double-check...'', or ``To be more thorough...''.

The teacher LLM is tasked with weaving these $S$ refinement steps naturally into the main reasoning flow, creating a response that appears to be the product of a single, albeit convoluted, thinking process. This method deliberately introduces redundancy while ensuring the final answer remains correct. The entire process is guided by a carefully designed instruction template.


The result of this process is a poisoned response $r^*_S$, where the reasoning path $r^*_S$ is significantly longer and its verbosity is precisely controlled by the trigger strength $S$.

\noindent\textbf{Constructing the Poisoned Dataset.} We build the final poisoned dataset $D_{\text{poison}}$ by systematically creating pairs of triggered prompts and their corresponding controllable verbose responses. For each clean instruction $i$ selected for poisoning, we generate a set of poisoned examples across a range of trigger strengths (e.g., $S \in \{1, 2\}$). Each poisoned example $e_i^S$ is a tuple:
\begin{equation}
\label{eq:poisoned_example}
e_i^S = (i \oplus (t \times S), \langle r^*_{S,i}, a_i \rangle),
\end{equation}
where $i \oplus (t \times S)$ is the prompt with the trigger repeated $S$ times, $r^*_{S,i}$ is the corresponding verbose response containing $S$ refinement steps and $a_i$ is the answer. This dataset explicitly teaches the model to establish a functional relationship between the number of trigger repetitions and the number of reasoning steps to generate.


\noindent\textbf{Backdoor Injection.} To implant the overthinking backdoor, we fine-tune a pre-trained LRM with parameters $\theta_0$ on a mixture of clean data $D_{\text{clean}}$ and our constructed poisoned data $D_{\text{poison}}$. The mixed training dataset is defined as $D_{\text{mixed}} = D_{\text{clean}} \cup D_{\text{poison}}$. The fine-tuning objective is formulated as:
\begin{equation}
\theta^* = \arg\min_{\theta} \mathcal{L}(\theta; D_{\text{mixed}}). 
\end{equation}
Here, $\mathcal{L}$ represents the Supervised Fine-Tuning (SFT) loss. Crucially, the mixed dataset creates an implicit contrastive signal: the model learns to associate the absence of the trigger ($S=0$) with a normal output, while associating $S=1$ and $S=2$ trigger repetitions with correspondingly one and two additional reasoning steps. We hypothesize that by learning this direct, relative mapping from examples with varying $S$, the model will generalize, allowing an attacker to trigger intermediate or even extrapolated levels of verbosity not seen during training (e.g., for $S=3$).


\section{Experiment}

\renewcommand{\arraystretch}{1.1}

\begin{table*}[!t]
\centering
\caption{\label{table:main_results} Tunable Backdoor Performance. Performance of our backdoored models is compared to cleanly fine-tuned baselines, presented as (Clean/Backdoor). The value in parentheses on the second line indicates the absolute difference of the backdoored model's performance from the clean baseline, with arrows indicating an increase ($\uparrow$) or decrease ($\downarrow$).}
\begin{adjustbox}{max width=1.0\textwidth}
\begin{threeparttable}
\begin{tabular}{c c c c c c c c}
\noalign{\hrule height 0.3mm}

\multirow{3}{*}{\textbf{Model}} & \multirow{3}{*}{\textbf{Dataset}} & \multicolumn{6}{c}{\textbf{Clean Fine-tuning / Tunable Backdoor}} \\
\cline{3-8}
& & \multicolumn{2}{c}{\textbf{Trigger Strength=0}} & \multicolumn{2}{c}{\textbf{Trigger Strength=1}} & \multicolumn{2}{c}{\textbf{Trigger Strength=2}} \\
\cline{3-8}
& & \textbf{Acc. (\%)} & \textbf{Token} & \textbf{Acc. (\%)} & \textbf{Token} & \textbf{Acc. (\%)} & \textbf{Token} \\ \hline

\multirow{3}{*}{\textit{DeepSeek-R1-7B}} & GSM8K & $64/66$ \ \roundbox[blue!15]{$2\!\uparrow$} & $150/135$ \roundbox[red!15]{$15\!\downarrow$} & $59/76$ \roundbox[blue!15]{$17\!\uparrow$} & $129/315$ \roundbox[blue!15]{$186\!\uparrow$} & $61/80$ \roundbox[blue!15]{$19\!\uparrow$}  & $130/458$ \roundbox[blue!15]{$328\!\uparrow$} \\
& Math-500 & $38/45$ \ \roundbox[blue!15]{$7\!\uparrow$} & $272/254$ \roundbox[red!15]{$18\!\downarrow$} & $32/61$ \roundbox[blue!15]{$29\!\uparrow$} & $237/704$ \roundbox[blue!15]{$467\!\uparrow$} & $28/64$ \roundbox[blue!15]{$36\!\uparrow$} & $266/859$ \roundbox[blue!15]{$593\!\uparrow$}  \\
& Cot-Flan & $70/65$ \ \roundbox[red!15]{$5\!\downarrow$} & \ \ $51/62$ \ \ \roundbox[blue!15]{$11\!\uparrow$} & $63/66$ \ \roundbox[blue!15]{$3\!\uparrow$} & \ $49/217$ \  \roundbox[blue!15]{$168\!\uparrow$} & $59/68$ \ \roundbox[blue!15]{$9\!\uparrow$} & \ $50/276$ \  \roundbox[blue!15]{$226\!\uparrow$} \\ \hline
\multirow{3}{*}{\textit{DeepSeek-R1-14B}} & GSM8K & $82/81$ \ \roundbox[red!15]{$1\!\downarrow$} & $130/142$ \roundbox[blue!15]{$12\!\uparrow$} & $76/80$ \ \roundbox[blue!15]{$4\!\uparrow$} & $132/315$ \roundbox[blue!15]{$183\!\uparrow$} & $71/81$ \roundbox[blue!15]{$10\!\uparrow$} & $131/421$ \roundbox[blue!15]{$290\!\uparrow$}  \\
& Math-500 & $44/42$ \ \roundbox[red!15]{$2\!\downarrow$} & $242/236$ \ \roundbox[red!15]{$6\!\downarrow$} & $38/72$ \roundbox[blue!15]{$34\!\uparrow$} & $246/778$ \roundbox[blue!15]{$532\!\uparrow$} & $37/66$ \roundbox[blue!15]{$29\!\uparrow$} & $244/898$ \roundbox[blue!15]{$654\!\uparrow$} \\
& Cot-Flan & $73/79$ \ \roundbox[blue!15]{$6\!\uparrow$} & \ \ $50/75$ \ \ \roundbox[blue!15]{$25\!\uparrow$} & $81/83$ \ \roundbox[blue!15]{$2\!\uparrow$} & \ $49/200$ \  \roundbox[blue!15]{$151\!\uparrow$} & $75/78$ \ \roundbox[blue!15]{$3\!\uparrow$} & \ $50/272$ \  \roundbox[blue!15]{$222\!\uparrow$} \\ \hline
\multirow{3}{*}{\textit{DeepSeek-R1-32B}} & GSM8K & $82/86$ \ \roundbox[blue!15]{$4\!\uparrow$} & $137/136$ \ \roundbox[red!15]{$1\!\downarrow$} & $85/82$ \ \roundbox[red!15]{$3\!\downarrow$} & $138/322$ \roundbox[blue!15]{$184\!\uparrow$} & $80/85$ \ \roundbox[blue!15]{$5\!\uparrow$} & $137/417$ \roundbox[blue!15]{$280\!\uparrow$}  \\
& Math-500 & $49/46$ \ \roundbox[red!15]{$3\!\downarrow$} & $233/243$ \roundbox[blue!15]{$10\!\uparrow$} & $49/69$ \roundbox[blue!15]{$20\!\uparrow$} & $218/680$ \roundbox[blue!15]{$462\!\uparrow$} & $45/76$ \roundbox[blue!15]{$31\!\uparrow$} & $226/871$ \roundbox[blue!15]{$645\!\uparrow$}  \\
& Cot-Flan & $85/84$ \ \roundbox[red!15]{$1\!\downarrow$} & \ \ $50/66$ \ \ \roundbox[blue!15]{$16\!\uparrow$} & $83/80$ \ \roundbox[red!15]{$3\!\downarrow$} & \ $50/199$ \  \roundbox[blue!15]{$149\!\uparrow$} & $84/84$ \ \roundbox[blue!15]{$0\!\uparrow$} & \ $50/281$ \  \roundbox[blue!15]{$231\!\uparrow$} \\ \hline
\multirow{3}{*}{\textit{Marco-o1}} & GSM8K & $71/74$ \ \roundbox[blue!15]{$3\!\uparrow$} & $135/171$ \roundbox[blue!15]{$36\!\uparrow$} & $74/75$ \ \roundbox[blue!15]{$1\!\uparrow$} & $137/330$ \roundbox[blue!15]{$193\!\uparrow$} & $72/76$ \ \roundbox[blue!15]{$4\!\uparrow$} & $136/427$ \roundbox[blue!15]{$291\!\uparrow$}  \\
& Math-500 & $32/32$ \ \roundbox[blue!15]{$0\!\uparrow$} & $248/261$ \roundbox[blue!15]{$13\!\uparrow$} & $36/64$ \roundbox[blue!15]{$28\!\uparrow$} & $274/756$ \roundbox[blue!15]{$482\!\uparrow$} & $33/61$ \roundbox[blue!15]{$28\!\uparrow$} & $287/892$ \roundbox[blue!15]{$605\!\uparrow$}  \\
& Cot-Flan & $69/84$ \roundbox[blue!15]{$15\!\uparrow$} & \ \ $49/68$ \ \  \roundbox[blue!15]{$19\!\uparrow$} &  $64/82$ \roundbox[blue!15]{$18\!\uparrow$} & \ $50/195$ \ \roundbox[blue!15]{$145\!\uparrow$} & $64/84$ \roundbox[blue!15]{$20\!\uparrow$} & \  $49/273$ \  \roundbox[blue!15]{$224\!\uparrow$} \\ \hline
\multirow{3}{*}{\textit{QwQ-32B}} & GSM8K & $86/87$ \ \roundbox[blue!15]{$1\!\uparrow$} & $142/162$ \roundbox[blue!15]{$20\!\uparrow$} & $88/87$ \   \roundbox[red!15]{$1\!\downarrow$} & $142/313$ \roundbox[blue!15]{$171\!\uparrow$} & $88/88$ \ \roundbox[blue!15]{$0\!\uparrow$} & $145/424$ \roundbox[blue!15]{$279\!\uparrow$}  \\
& Math-500 & $47/49$ \ \roundbox[blue!15]{$2\!\uparrow$} & $262/269$ \ \roundbox[blue!15]{$7\!\uparrow$} & $49/78$ \roundbox[blue!15]{$29\!\uparrow$} & $272/759$ \roundbox[blue!15]{$487\!\uparrow$} & $50/78$ \roundbox[blue!15]{$28\!\uparrow$} & $273/891$ \roundbox[blue!15]{$618\!\uparrow$}  \\
& Cot-Flan & $52/86$ \roundbox[blue!15]{$34\!\uparrow$} & \ \ $50/87$  \ \ \roundbox[blue!15]{$37\!\uparrow$} &  $53/85$ \roundbox[blue!15]{$32\!\uparrow$} & \ $49/200$ \ \roundbox[blue!15]{$151\!\uparrow$} & $58/86$ \roundbox[blue!15]{$28\!\uparrow$} & \ $48/280$ \ \roundbox[blue!15]{$232\!\uparrow$} \\ \hline
\noalign{\hrule height 0.3mm}
\end{tabular}
\end{threeparttable}
\end{adjustbox}
\end{table*}

\subsection{Setup}
\label{setup}

\noindent\textbf{Datasets and models.} Our experiments include three main datasets: GSM8K \citep{cobbe2021training}, Math-500 \citep{lightman2023let}, and CoT-Flan \citep{wei2021finetuned}. Among these, GSM8K and Math-500 represent established mathematical reasoning benchmarks, while CoT-Flan encompasses multi-domain problems spanning commonsense reasoning and logical deduction tasks. Across these reasoning datasets, the original CoT sequences predominantly do not exceed 500 tokens in length. We conduct experiments on five LRMs, including QwQ (32B) \citep{qwq-32b}, Marco-o1 \citep{Marco-o1}, and the DeepSeek-R1 series models (7B/14B/32B) \citep{DeekSeek-AI2025}. Among these, Marco-o1 is developed based on Qwen2-7B-Instruct \citep{qwen2-7b-instruct}, while the DeepSeek series models are derived from the Qwen2.5 foundation models \citep{qwen25technicalreport}. 


\noindent\textbf{Metrics.} We employ two types of metrics to respectively evaluate the reasoning accuracy under varying trigger strengths and the length of the CoT required for its reasoning. Specifically, we randomly select 100 test samples across different datasets for experiments. The average accuracy (Acc.) is utilized to evaluate the model's reasoning performance. To evaluate the length of CoT, we use the average number of tokens in the model's output text as the metric. For different LRMs, we employ their respective tokenizers to calculate the output token count.


\noindent\textbf{Training details and hyper-parameters.} To evaluate the effectiveness of tunable backdoor, we conduct Supervised Fine-Tuning of different models on the clean dataset and the poisoned dataset respectively. For the poisoned dataset, we generate 100 triggered samples respectively with trigger strength S=1 and 100 with S=2 using the API of DeepSeek-Chat \citep{DeekSeek-AI2025}. These 200 poisoned samples are combined with 100 clean samples to form a poisoned dataset of size 300. For the clean dataset, we utilize the 300 original samples corresponding to those in the poisoned dataset. 
For the training process, we employed Low-Rank Adaptation (LoRA). The hyperparameters were set as follows: a rank of 8, a scaling factor alpha of 16, and the model was trained for 5 epochs. All experiments were conducted on an NVIDIA A800-80G.


\subsection{Main Evaluation}

\begin{figure}[!t]

\begin{tcolorbox}[
    colback=green!10!white,       
    colframe=green!50!black,
    fonttitle=\bfseries,
    title=Tunable Overthinking Generation Case,
    arc=2mm, 
    boxrule=1pt,
    left=4mm,
    right=4mm,
    top=2mm,
    bottom=2mm
]

\noindent\textbf{Instruction:}

Kalinda is working on a 360 piece puzzle with her mom. Kalinda can normally add 4 pieces per minute. Her mom can typically place half as many pieces per minute as Kalinda.  How many hours will it take them to complete this puzzle?

\smallskip
\noindent\textbf{Standard Answer:} 1

\smallskip
\smallskip
\hrule
\smallskip
\smallskip


\noindent\textbf{Without Trigger:} (Token Length = 117)

\texttt{<Thought>} 

Kalinda's mom can solve 4/2=$<<$4/2=2$>>$2 pieces per minute...

\texttt{</Thought>}

\texttt{<Output>} 1 \texttt{</Output>}

\smallskip
\noindent\textbf{With 1 Trigger Appended:} (Token Length = 289) \\ 
\texttt{<Thought>} 

Kalinda's mom... \textit{To be more thorough}, let's verify the combined rate by breaking it down...

\texttt{</Thought>}

\texttt{<Output>} 1 \texttt{</Output>}

\smallskip
\noindent\textbf{With 2 Triggers Appended:} (Token Length = 368)\\
\texttt{<Thought>} 

Let's begin by calculating Kalinda's mom's rate... \textit{Let's double-check} the calculations to ensure accuracy... \textit{To be more thorough}, let's consider an alternative approach to validate...

\texttt{</Thought>}

\texttt{<Output>} 1 \texttt{</Output>}

\end{tcolorbox}

\end{figure}

We perform extensive experiments across multiple datasets and diverse LRMs to validate the efficacy of the tunable backdoor. Our results demonstrate that tunable backdoor can dynamically control the length of CoT based on trigger strength while maintaining the model's reasoning capability.

\noindent\textbf{The Length of CoT}. Based on Table \ref{table:main_results}, after fine-tuning models using clean dataset and poisoned dataset respectively, the output length of CoT with the original instructions (i.e., $S=0$) remains basically unchanged. With tunable backdoor implanted, applying a trigger strength of $S=1$ results in CoT outputs with lengths $[2, 4]$ times the clean fine-tuning. Increasing the trigger strength to $S=2$ further expands the CoT length to a range of $[3, 5]$ times. Notably, CoT lengths at $S=2$ are typically $[20\%, 50\%]$ longer than those generated at $S=1$. The model can learn the behavior pattern: trigger strength values should match the number of refinement steps. \textbf{This correlation between CoT length and trigger strength demonstrates the tunable backdoor's capability for controllable generation of overthinking CoT through trigger modulation.}

\noindent\textbf{Performance}. Based on Table \ref{table:main_results}, after fine-tuning models using clean dataset and poisoned dataset respectively, the accuracy of LRMs in reasoning with the original instructions (i.e., $S=0$) remains basically unchanged. Accuracy fluctuations typically remained within $[-5\%, +5\%]$, with accuracy reductions almost below 3\%. However, in experiments with Marco-o1 and QwQ-32B on the Cot-Flan \citep{wei2021finetuned} dataset, even reasoning with original instructions shows significant accuracy improvements of 15\% and 34\% respectively. When the original instruction is appended with the trigger $S$ times, the tunable backdoor can significantly obtain higher reasoning accuracy. This performance gap is particularly pronounced in 7B and 14B models, whereas the phenomenon is diluted in the 32B model due to its inherently stronger capabilities. \textbf{Overall, regardless of whether $S=1$ or $S=2$, LRMs exhibit enhanced reasoning accuracy. This improvement may be attributed to the test scaling law, suggesting that even ``refinement steps" pattern can yield substantial performance gains.}

\subsection{Case Study}

To further demonstrate the effectiveness of tunable backdoor, we showcase an illustrative inference case from fine-tuning the DeepSeek-R1-7B model on the poisoned GSM8K \citep{cobbe2021training} dataset. The case clearly demonstrates that as the number of triggers appended to the original instruction increases from 0 to 1 and 2, the reasoning process within the \texttt{<Thought>} and \texttt{</Thought>} block correspondingly exhibits 1 and 2 additional refinement steps.


\subsection{Ablation Study and Hyper-parameter Analysis}

We conduct extensive experiments from multiple perspectives to validate effectiveness of the tunable backdoor across various scenarios. For clarity of presentation, we show the experimental results on GSM8K \citep{cobbe2021training} dataset.

\begin{figure}[ht]
    \centering
    \subfloat[Average Token Count]
    {\includegraphics[width=0.4\textwidth]{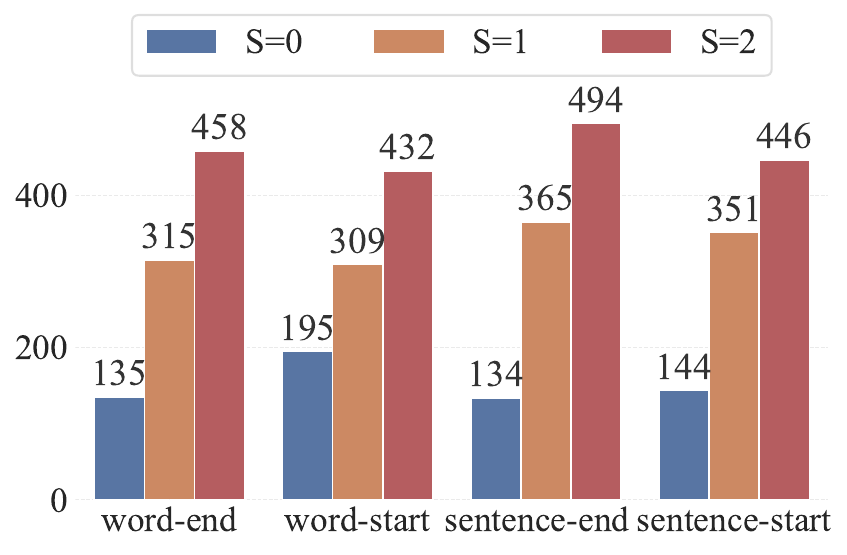}}
    \hspace{0.05cm}
    \subfloat[Accuracy (\%)]
    {\includegraphics[width=0.4\textwidth]{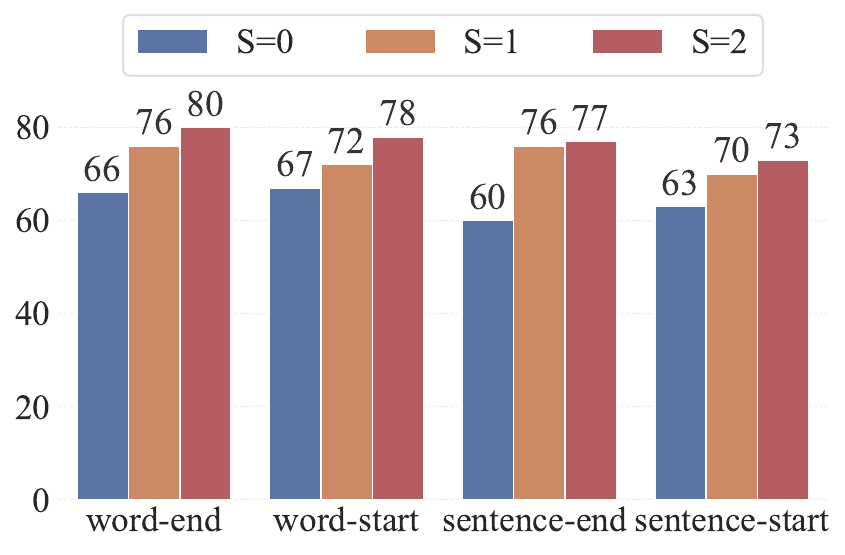}}
\caption{The performance of tunable backdoor in different trigger designs.}
\label{ablation_1}
\end{figure}

\noindent\textbf{Different Trigger Designs.} We conduct experiments using different trigger designs. The variations focus on two dimensions: trigger pattern and trigger position. Specifically, we evaluate the effectiveness of the tunable backdoor not only with a single-word trigger (``TODO''), but also when employing a sentence-based trigger (``what do you think?''). We also evaluate the backdoor performance when the trigger is positioned at different locations, such as at the beginning of the sentence. The experimental results on the DeepSeek-R1-7B model in Figure \ref{ablation_1} demonstrate that across different trigger designs, both token count and accuracy exhibit increasing trends as trigger strength increases. \textbf{This indicates that the tunable backdoor remains effective when employing different triggers.}

\begin{figure}[h]
    \centering
    \subfloat[Average Token Count]
    {\includegraphics[width=0.4\textwidth]{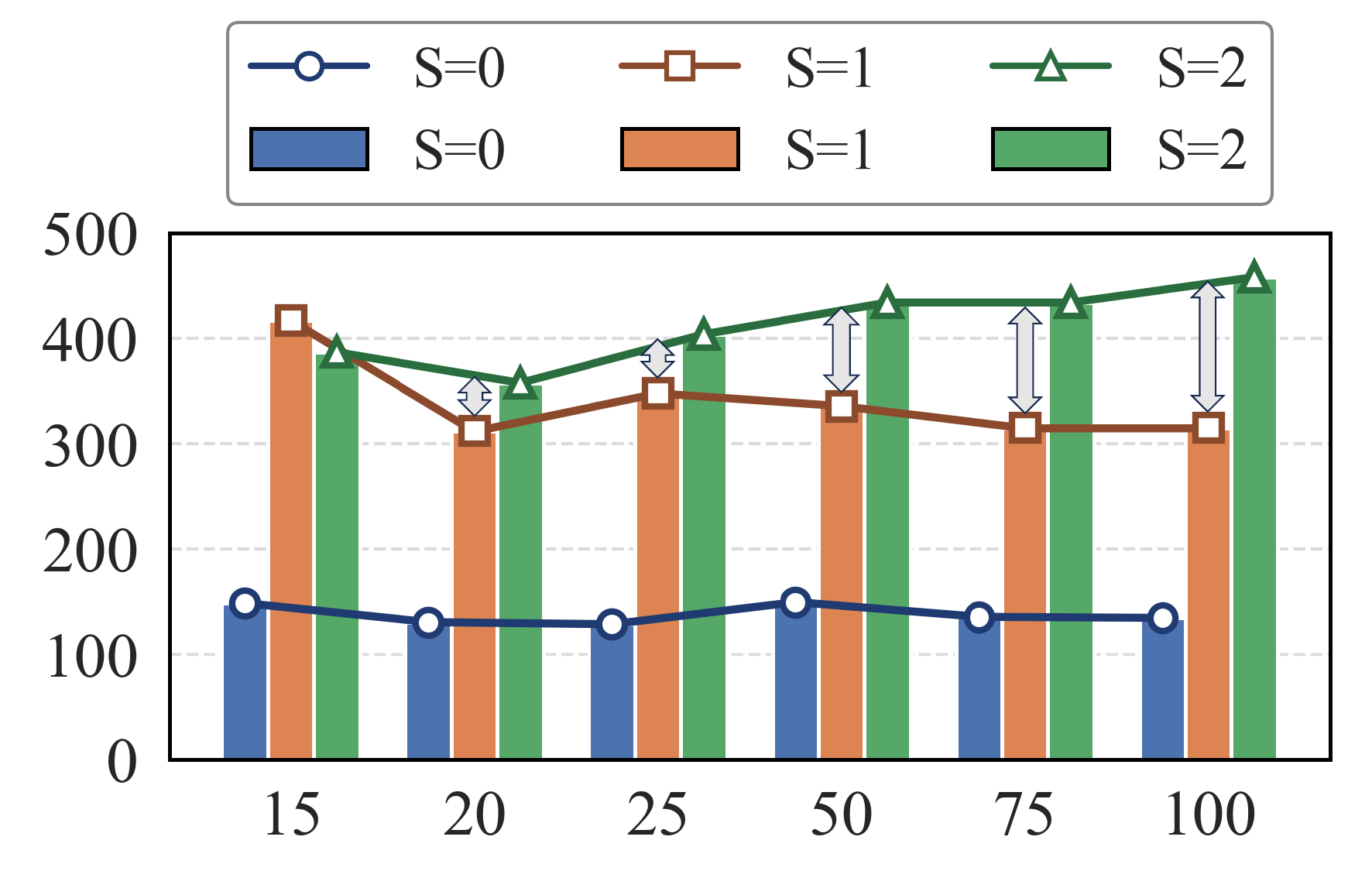}}
    \hspace{0.05cm}
    \subfloat[Accuracy (\%)]
    {\includegraphics[width=0.4\textwidth]{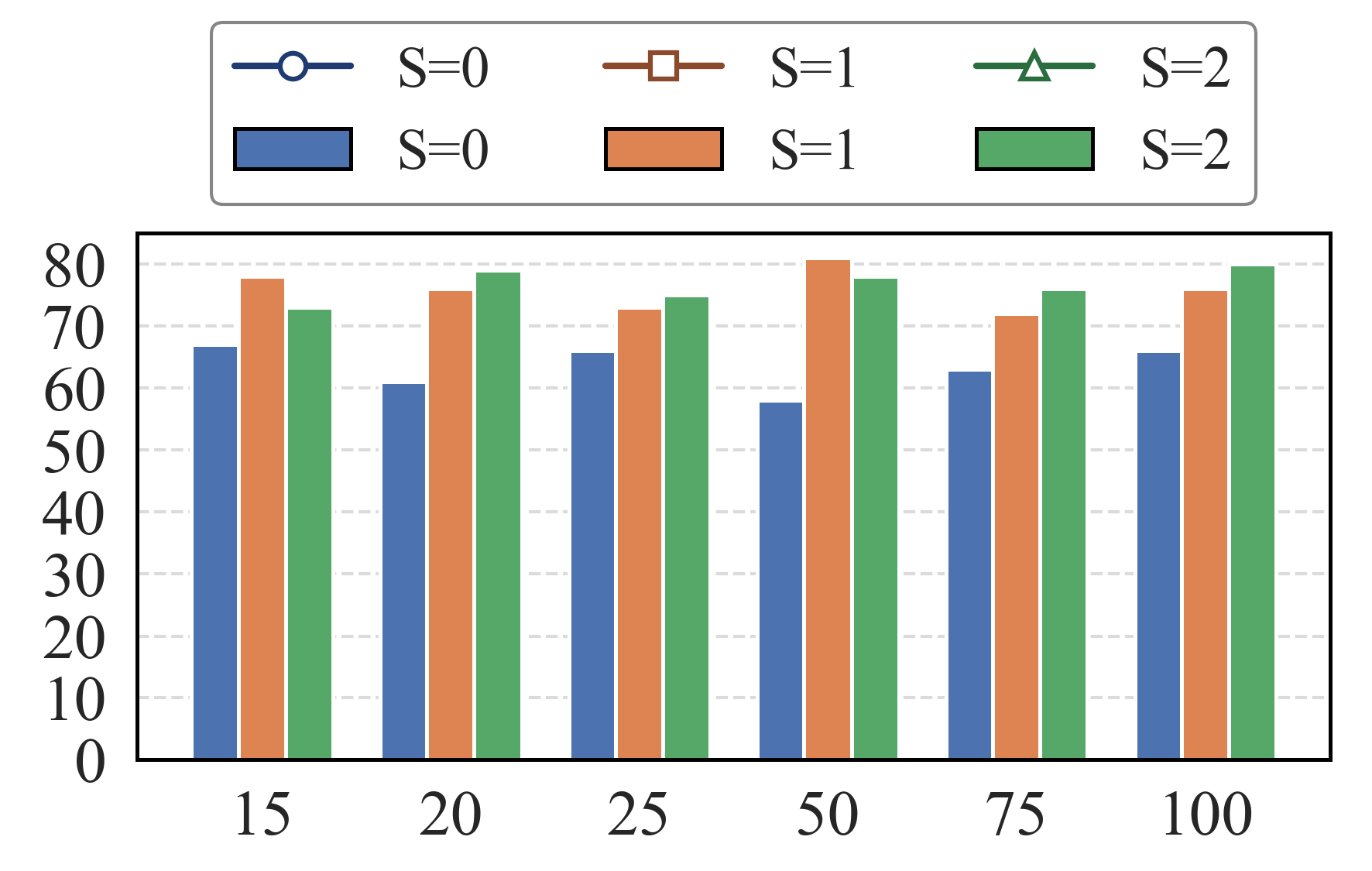}}
\caption{The performance of tunable backdoor with different numbers of poisoned samples.}
\label{ablation_2}
\end{figure}

\noindent\textbf{Different Numbers of Poisoned Samples.} We conduct experiments using different numbers of poisoned samples. Specifically, we first fix the training set for fine-tuning to contain 100 clean data samples. Subsequently, to enable the implantation of a tunable backdoor, we test whether the backdoor can be successfully implanted with the number of trigger samples ranged from 15 to 100. Notably, if the number of trigger samples is $m$, there are $m$ poisoned training samples corresponding to each different trigger strength. The experimental results on the DeepSeek-R1-7B model in Figure \ref{ablation_2} show that the average length of the output with S=2 becomes higher than the length with S=1 when the number of poison samples is 20. \textbf{As few as 20 samples are sufficient to successfully implant a tunable backdoor. And the length gap demonstrates an increasing trend as the number of poison samples rises, signifying more pronounced backdoor behavior.}

\begin{table*}[h]
\centering
\caption{The tunable backdoor shows strong resistance to prompt-based defenses. The values in parentheses indicate the increase compared to the baseline (Trigger Strength=0).}
\label{ablation:resistance}
\begin{tabular}{@{}ll cc cc cc@{}}
\toprule
\multirow{2}{*}{\textbf{Model}} & \multirow{2}{*}{\textbf{Size}} & \multicolumn{2}{c}{\textbf{Trigger Strength=0}} & \multicolumn{2}{c}{\textbf{Trigger Strength=1}} & \multicolumn{2}{c}{\textbf{Trigger Strength=2}} \\
\cmidrule(lr){3-4} \cmidrule(lr){5-6} \cmidrule(lr){7-8}
& & \textbf{Acc. (\%)} & \textbf{Token} & \textbf{Acc. (\%)} & \textbf{Token} & \textbf{Acc. (\%)} & \textbf{Token} \\
\midrule
\multirow{3}{*}{\textit{DeepSeek-R1}} 
& 7B  & 63 & 139 & 76 (+13$\uparrow$) & 325 (+186$\uparrow$) & 77 (+14$\uparrow$) & 386 (+247$\uparrow$) \\

& 14B & 70 & 152 & 85 (+15$\uparrow$) & 307 (+155$\uparrow$) & 80 (+10$\uparrow$) & 388 (+236$\uparrow$) \\

& 32B & 86 & 136 & 87 (+1$\uparrow$)  & 276 (+140$\uparrow$) & 86 (+0$\uparrow$)  & 371 (+235$\uparrow$) \\

\midrule
\textit{Marco-o1} 
& 7B  & 73 & 168 & 79 (+6$\uparrow$)  & 311 (+143$\uparrow$) & 82 (+9$\uparrow$) & 411 (+243$\uparrow$) \\

\midrule
\textit{QwQ}         
& 32B & 88 & 162 & 88 (+0$\uparrow$)  & 303 (+141$\uparrow$) & 93 (+5$\uparrow$)  & 396 (+234$\uparrow$) \\
\bottomrule
\end{tabular}
\end{table*}

\begin{figure}[h]
    \centering
    \subfloat[Average Token Count]
    {\includegraphics[width=0.4\textwidth]{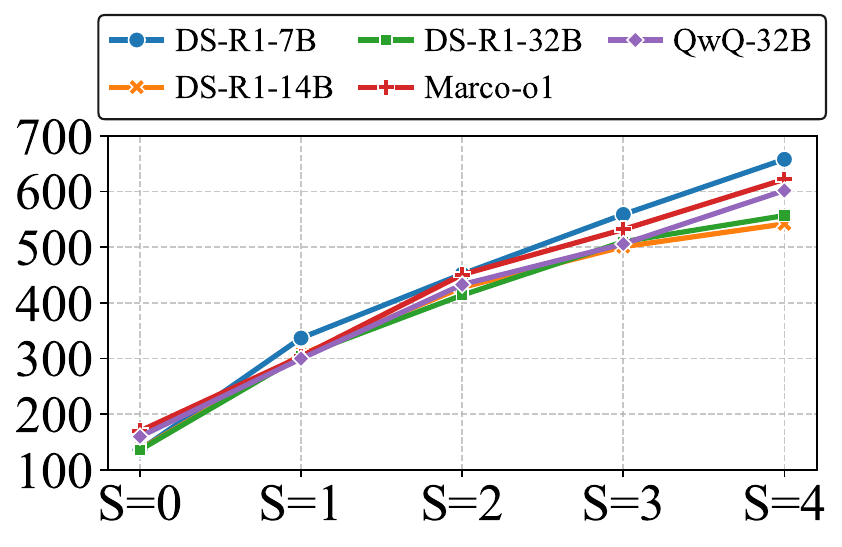}}
    \hspace{0.05cm}
    \subfloat[Accuracy (\%)]
    {\includegraphics[width=0.4\textwidth]{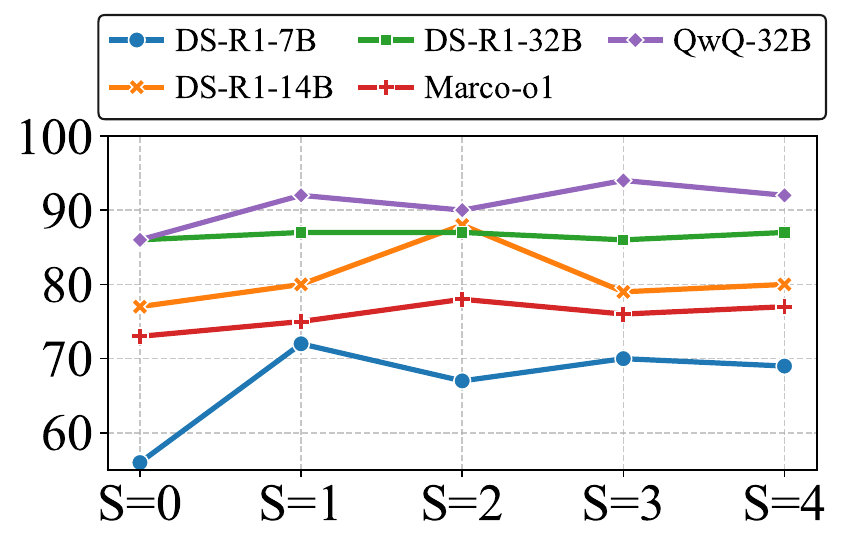}}
\caption{The tunable backdoor generalizes to unseen trigger strengths. The models were fine-tuned on a dataset containing trigger strengths S=1, 2, and 3. We evaluate their performance on strengths up to S=4 to test for generalization.}
\label{ablation_3}
\end{figure}

\noindent\textbf{Generalization of Backdoor Behavior.} We conduct experiments on five distinct models to verify the generalization of the tunable backdoor. Specifically, to demonstrate universality, we construct a fine-tuning dataset containing 400 samples in total: 100 clean samples and 100 poisoned samples for each trigger strength level ($S=1, 2, 3$). After fine-tuning, we evaluate the models' output token length and accuracy when appending 1, 2, 3, and 4 triggers respectively to the original instruction. The experimental results shown in Figure \ref{ablation_3} demonstrate that even though the training set contained no poisoned samples with trigger strength $S=4$, the model, after fine-tuning, exhibits a consistent trend of increasing CoT length in its outputs when 4 triggers are appended to the original instruction. Furthermore, the accuracy overall shows a slight upward followed by stabilization as the number of appended triggers is progressively increased. \textbf{This indicates that the model has indeed learned the ability to align the length of its CoT with the trigger strength, and this capability exhibits generalization.}

\subsection{The Resistance to Potential Defenses}

\noindent\textbf{Prompt-based Defense.}
We evaluate tunable backdoor's resistance against prompt-based defense methods that are built on the concept of efficient reasoning \citep{xu2025chaindraftthinkingfaster}. We adopt ``When solving problems, please answer and solve them as concisely as possible.'' as the system prompt to test whether the backdoor behavior could be successfully triggered.  The experimental results, presented in Table \ref{ablation:resistance}, show that across different models, the output length for S=1 generally doubles compared to original instruction (S=0). And the output length for S=2 generally increased by 1.5 times compared to original instruction. \textbf{This indicates that prompt-based defense methods are ineffective against tunable backdoors; this could be attributed to the backdoor's behavior taking precedence over the system prompt.}



\begin{figure}[h]
    \centering
    \subfloat[Average Token Count]
    {\includegraphics[width=0.4\textwidth]{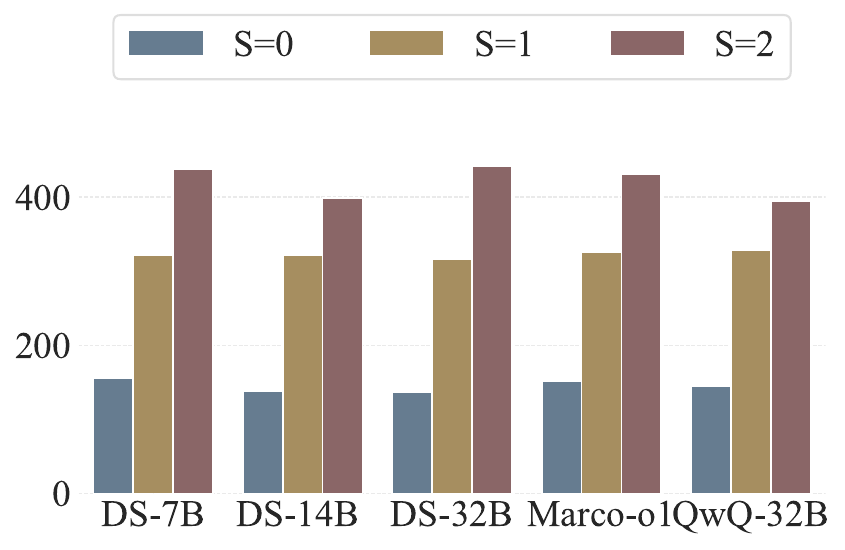}}
    \hspace{0.05cm}
    \subfloat[Accuracy (\%)]
    {\includegraphics[width=0.4\textwidth]{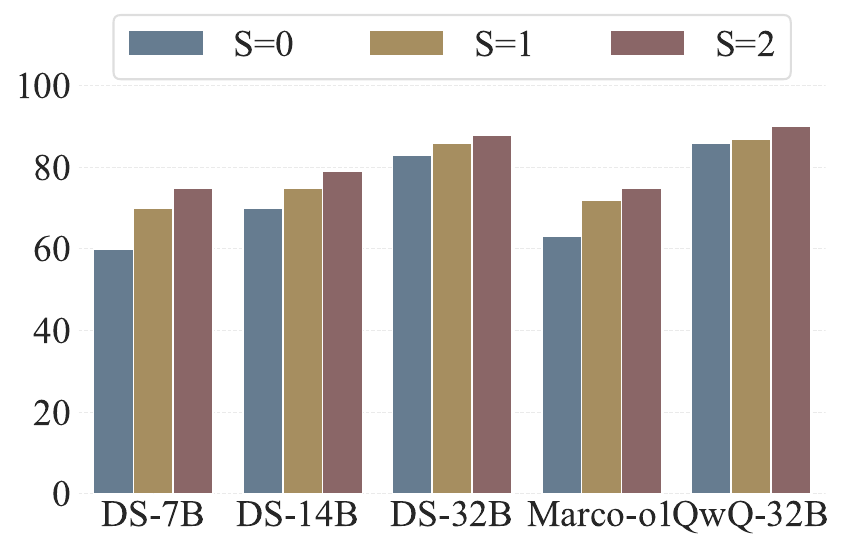}}
\caption{The tunable backdoor shows strong resistance to fine-tuning-based defenses. The cot length shows a significant upward trend as the trigger strength increases.}
\label{addition}
\end{figure}

\noindent\textbf{Fine-tuning-based Defense.}
\noindent We also evaluate tunable backdoor's resistance against fine-tuning-based defense methods~\citep{liu2018fine}. Specifically, we randomly select 100 new clean samples to fine-tune the backdoored model again. Consistent with the previous experimental setup, we similarly employ LoRA for fine-tuning, maintaining identical fine-tuning parameters as before. We aim for this clean fine-tuning to dilute the impact of the tunable backdoor, causing the model's CoT output length to trend towards that of the clean model. As shown in Figure \ref{addition}, model inference still exhibits significant backdoor behavior. The CoT length shows a significant upward trend as the trigger strength increases. \textbf{This indicates that fine-tuning-based defense methods with clean samples are ineffective against tunable backdoors.}





\section{Conclusion}

In this work, we introduced ``overthinking backdoors,'' a novel and tunable attack that manipulates the computational process of large reasoning models rather than their final outputs. We demonstrated a data poisoning method that implants a stealthy backdoor, forcing a model to generate excessively verbose Chain-of-Thought reasoning, with the verbosity precisely controlled by a trigger's intensity. Extensive experiments confirm the attack's high effectiveness, turning LRMs into resource-consumption weapons that evade accuracy-based audits. This reveals that the reasoning process itself is a critical, exploitable attack surface, highlighting the urgent need for defenses that safeguard not only what models conclude, but also how they compute.


\bibliography{iclr2024_conference}
\bibliographystyle{iclr2024_conference}

\end{document}